\pdfoutput=1

\PassOptionsToPackage{table}{xcolor}
\documentclass[11pt]{article}
\usepackage{ACL2023}

\usepackage{times}
\usepackage{latexsym}
\usepackage{amsmath}
\usepackage{amssymb}
\usepackage{array, multirow, bigdelim, makecell, booktabs} 
\usepackage{cleveref}
\usepackage{graphicx}
\usepackage{subcaption}
\usepackage{adjustbox}

\usepackage{comment}
\usepackage{float} 
\usepackage{amsmath}
\usepackage{algorithm}
\usepackage{algpseudocode}

\newcommand{\methodNameNoSpace}{IGBP}
\newcommand{\methodName}{IGBP }
\newcommand{\methodFullName}{Iterative Gradient-Based Projection }
\newcommand{\lossFuncName}{projective loss }
\newcommand{\lossFuncNameNoSpace}{projective loss}

\usepackage{amsmath}

\usepackage[T1]{fontenc}

\usepackage[utf8]{inputenc}

\usepackage{microtype}

\usepackage{inconsolata}

\title{Shielded Representations: Protecting Sensitive Attributes Through Iterative Gradient-Based Projection}

\author{Shadi Iskander \hspace{2em} Kira Radinsky \hspace{2em} Yonatan Belinkov\thanks{~~Supported by the Viterbi Fellowship in the Center for Computer Engineering at the Technion.} \\
  {\tt shadi.isk@campus.technion.ac.il} \\
  {\tt kirar@cs.technion.ac.il \hspace{2em} belinkov@technion.ac.il} \\
{Technion -- Israel Institute of Technology}}

\begin{document}

\maketitle

\newcommand{\indep}{\perp \!\!\! \perp}
\begin{abstract}
Natural language processing models tend to learn and encode social biases present in the data. One popular approach for addressing such biases is to eliminate encoded information from the model's representations. However, current methods are restricted to removing only linearly encoded information.
In this work, we propose \methodFullName (\methodNameNoSpace), a novel method for removing non-linear encoded concepts from neural representations. Our method consists of iteratively training neural classifiers to predict a particular attribute we seek to eliminate, followed by a projection of the  representation on a hypersurface, such that the classifiers become oblivious to the target attribute. We evaluate the effectiveness of our method on the task of removing gender and race information as sensitive attributes. Our results demonstrate that \methodName is effective in mitigating bias through intrinsic and extrinsic evaluations, with minimal impact on downstream task accuracy.\footnote{Code is available at \url{https://github.com/technion-cs-nlp/igbp_nonlinear-removal}.}
\end{abstract}
\section{Introduction}

The increasing reliance on natural language processing models in decision-making systems has led to a renewed focus on the potential biases that these models may encode. Recent studies have demonstrated that word embeddings exhibit gender bias in their associations of professions \cite{bolukbasi2016man,caliskan2017semantics} and that learned representations of language models capture demographic data about the writer of the text, such as race or age \cite{blodgett-etal-2016-demographic,DBLP:conf/emnlp/ElazarG18}. Model decisions can be affected by these encoded biases and irrelevant attributes, leading to a wide range of inequities toward certain demographics. For example, a model designed to review job resumes should not factor in the applicants' gender or race. Consequently, it is desirable to be able to manipulate the type of data encoded within text representations and to exclude any sensitive information in order to create more fair and equitable models.\\
\begin{figure}[t]
\includegraphics[width=0.48\textwidth,height=0.30\textwidth]{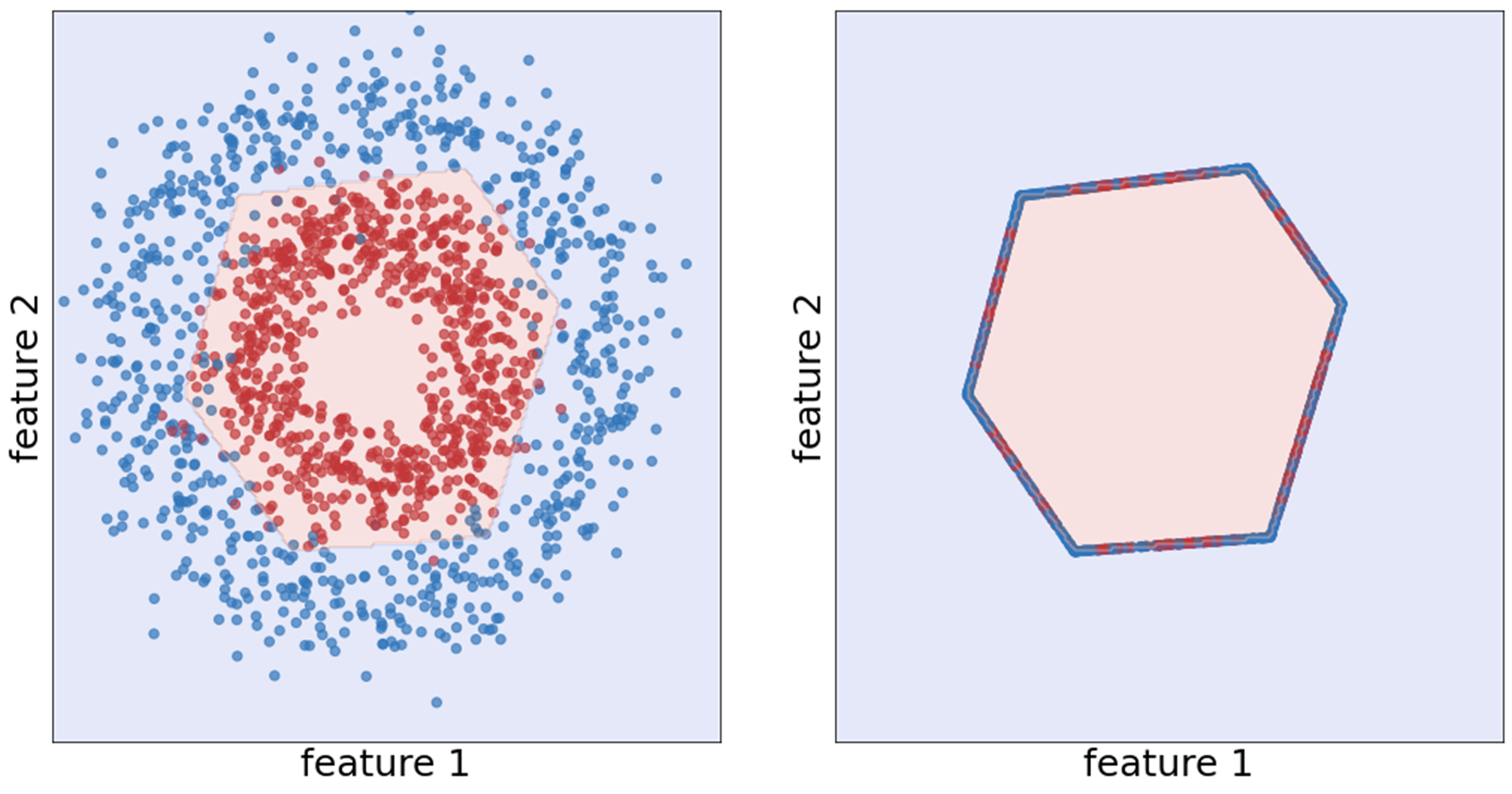}
\vspace{-0.5cm}
\caption{Left: Non-linear decision boundary of a ReLU 2-layer neural network on a binary classification task. Right: One iteration of \methodName algorithm produces clean, indistinguishable samples.}
\label{fig:nonlin-proj}
\end{figure} 
Removing the presence of sensitive attributes from the representations learned by deep neural networks is non-trivial, as these representations are often learned using complex and hard-to-interpret non-linear models. Re-training the language model can be a costly solution, therefore post-hoc removal methods that work at the representation layer have been proposed, such as linear projection of the embeddings on a hyperplane that distinguishes between the sensitive attribute \cite{bolukbasi2016man,DBLP:conf/acl/RavfogelEGTG20}. However, neural networks do not necessarily represent concepts in a linear manner. To address this issue, \citet{ravfogel-etal-2022-adversarial} proposed kernelization of a linear minimax game for concept erasure, but this approach is restricted to the selection of kernel and the attribute protection does not transfer to different types of non-linear probes. Accordingly, \citet{DBLP:conf/icml/RavfogelTGC22,ravfogel-etal-2022-adversarial} considered non-linear concept erasure to be an open problem.

In this paper, we propose a non-linear concept erasure method, \methodNameNoSpace, to eliminate information about the protected attribute from neural representations. 
We use a trained probe classifier that attempts to predict the protected attribute and a novel loss function suited for the task of concept removal.  Then, we leverage the gradients of this loss to guide for projection of the representations to a hypersurface that does not contain information used by the classifier regarding the sensitive-attribute. This is done by projecting the representations to the separating boundary of the classifier. Figure \ref{fig:nonlin-proj} illustrates a 2-dimensional example.

Our approach supports the use of non-linear neural classifiers. When used with a linear classifier, it is equivalent to Iterative Null Space Projection (INLP), a popular linear concept removal method \citep{DBLP:conf/acl/RavfogelEGTG20}.

We perform an empirical  evaluation of the proposed method using: (1) intrinsic evaluation of word embeddings measuring word-level gender bias removal and (2) extrinsic fair classification evaluation over tasks that uses contextualized word representations. The empirical results show that the proposed method is successful in sensitive-attribute removal and mitigating bias, outperforming competing algorithms with minimal impact on the downstream task accuracy.

\section{Related Work}
Many studies \citep[e.g.,][] {caliskan2017semantics,DBLP:conf/naacl/RudingerNLD18} investigated social biases in word embeddings and text representations. Recent work have showed how applications that use pre-trained representations reflect and amplify these kinds of social biases \cite{DBLP:conf/naacl/ZhaoWYOC18,DBLP:conf/emnlp/ElazarG18}.

The approaches tackling this problem can be categorized into three lines of work: pre-processing methods which manipulate the input distribution before training \citep[e.g.,][]{DBLP:conf/naacl/ZhaoWYOC18,DBLP:conf/iccv/WangZYCO19}, in-processing methods which focus on learning fair models during training \citep[e.g.,][]{xie2017controllable,DBLP:journals/corr/BeutelCZC17,zhang2018mitigating,DBLP:journals/corr/abs-2212-10563} and post-hoc methods \citep[e.g.,][]{DBLP:conf/acl/RavfogelEGTG20,wang-etal-2020-double, DBLP:conf/icml/RavfogelTGC22,ravfogel-etal-2022-adversarial}, which assume a fixed, pre-trained set of representations from any encoder and aim to learn a new set of unbiased representations. 

Since re-training a model can be costly, a lot of focus was given to post-hoc methods, which is the main focus of this work.

The most common post-hoc approach to remove sensitive information from word embeddings is to use a linear projection. \citet{bolukbasi2016man} identified a gender subspace, which is a subspace spanned by the directions of embeddings that capture the bias, such as the direction ‘‘he’’ -- ‘‘she’’. They suggested projecting all the gender-neutral word embeddings on the gender subspace's first principle component to make  neutral words equally distant from male and female-gendered words. However, \citet{DBLP:conf/acl-wnlp/GonenG19} showed that this method only covers up bias and not fully removes it from the representation. Another critical drawback of the method is that it requires user selection of a few gender directions.

\citet{DBLP:conf/acl/RavfogelEGTG20} tried to overcome this drawback of manually defining gender direction, and presented the Iterative Null-space Projection (INLP) method. It is based on training linear classifiers that predict the attribute they wish to remove, then projecting the representations on the classifiers' null-space.
\citet{DBLP:conf/icml/RavfogelTGC22} aims to linearly remove information from neural representations by using a linear minimax game-based approach, and derive a closed-form solution for certain objectives.
One of the limitations of linear removal methods is their inability to remove non-linear information about the protected attribute, which is often encoded in text representations through complex neural networks. In contrast, our method is capable of removing both linear and non-linear information, resulting in a more effective reduction of extrinsic bias (Section \ref{sec:eval_extrinsic}).

\citet{ravfogel-etal-2022-adversarial} proposed a nonlinear extension of the concept-removal objective of \citet{DBLP:conf/icml/RavfogelTGC22}. They identify the subspace to be neutralized in kernel space by running a kernelized version of a minimax game as in \citet{DBLP:conf/icml/RavfogelTGC22}. \citet{shao-etal-2023-gold} also use kernels to try and remove non-linear information
While this approach aims to remove non-linear information, it can only choose data mapping from a pre-defined set of kernels, and as shown in \citet{DBLP:conf/icml/RavfogelTGC22}, the attribute protection does not transfer to other non-linear kernels . Our approach uses a deep neural network as the bias signal modeling, thus has the potential to express any non-linear function. Our empirical results (Section \ref{sec:experiments}) show that our method significantly outperforms these methods on a variety of tasks. 

\section{Approach}

\subsection{Problem Formulation}
Given a dataset $D = \{x_i, y_i, z_i\}_{i=1}^N$ which consists of triples of text representation $x_i \in \mathcal{X}$, downstream task label  $y_i \in \mathcal{Y}$ and a protected attribute $z_i \in \mathcal{Z}$ which corresponds to discrete attribute values, such as gender. Our goal is to eliminate the information related to the protected attribute from the representations while minimizing the effect on other relevant information. To achieve this, we intend to learn a non-linear transformation of the representations such that the protected attribute $z_i$ cannot be inferred from the transformed representations $x^{clean}_i$, while still preserving the information with regard to the downstream task label $y_i$.

\subsection{Adversarial Approach Background}

The core of our approach is to produce projection of the representations such that any classifier is unable to distinguish between the protected attribute groups.
To gain some intuition about how such projections are generated, let us first consider a trained probe classifier $f$ that classifies the attribute label $z$ of each representation vector $x$. By assigning adversarial perturbations and moving in the direction of the gradient of the loss function with respect to the input vector, the representations can be modified such that the classifier's ability to predict the protected attribute is hindered, while minimizing the alteration of other relevant information:
\begin{equation}
x_{new} = x + \lambda\cdot\nabla_{x}L(f(x), z) ,
\label{eq:AdvCE}
\end{equation}
where $\lambda > 0$. \citet{DBLP:conf/emnlp/ElazarG18} applied a similar approach in removal of demographic attributes from text data during training. In contrast, we apply our method on the representation layer post-training, with a specific loss function and $\lambda$.

\begin{figure}
\includegraphics[width=0.48\textwidth,height=0.3\textwidth]{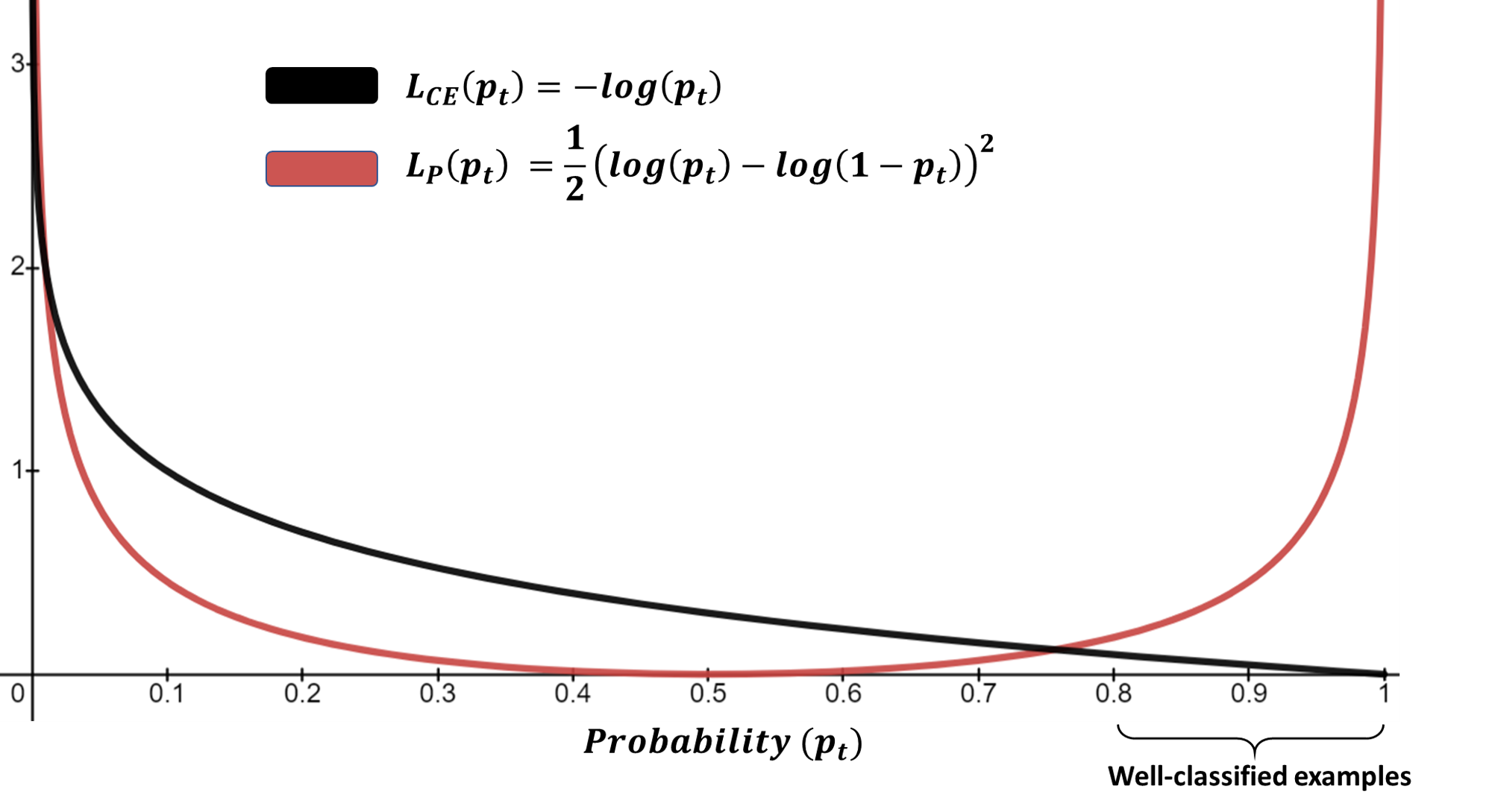}
\vspace{-0.5cm}
\caption{Cross entropy loss (black) and the \lossFuncName (red) behavior as a function of the probability of the probe classifier given an embedding.
The \lossFuncName emphasizes the well-classified, biased embeddings.} 
\label{fig:lossfunc}
\end{figure}

We present a novel loss for $L$, to which we call the \lossFuncNameNoSpace. It is designed for removing information from neural representation. It allows for a single-step projection of the representations, rendering the probe classifier $f$ oblivious to the protected attribute.
Before presenting the \lossFuncNameNoSpace, we explore why the common cross entropy (CE) is not optimal for our task.
The CE loss function is defined as:
\begin{equation}
L_{CE}(p,y) = \begin{cases}
  -log(p)  &   \text{if y=1} \\
  -log(1-p) &  \text{otherwise.}
\end{cases}
\end{equation}
where 
$y \in \{\pm1\}$ specifies the ground-truth class and
$p \in [0, 1]$ is the model’s estimated probability for the class
with label $y = 1$. For the sake of clarity, we formally define $p_t$:
\begin{equation}
p_t = \begin{cases}
  p  &   \text{if y=1} \\
  1-p &  \text{otherwise.}
\end{cases}
\end{equation}
and rewrite $CE(p,y) = CE(p_t) = -log(p_t)$.\\
The CE loss can be seen in black in Figure \ref{fig:lossfunc}. A noteworthy characteristic of this loss is that examples which are considered to have a strong signal of the protected attribute (i.e., are easily classified with $p_t \gg 0.5$) yield low gradients. In Appendix~\ref{app:math_grad_CE} we demonstrate mathematically that:
\begin{equation}
        \nabla_x L_{CE} =  \pm \: (1-p_t)\nabla_x f^\top\\
\label{eq:gradientsCE}
\end{equation}
As $p_t$ approaches 1, $\nabla_x L_{CE}$ tends to $0$ and the adversarial perturbation associated with the most-biased samples is vanishingly small.
Hence, the use of gradients of the CE loss for information removal brings about a major disadvantage.

\subsection{Projective Loss}
\label{sec:projloss}

A more effective way to remove the entire signal of bias in the representations would be projecting them on the hypersurface where the classifier is oblivious to the protected attribute.
To achieve this, we propose the \lossFuncNameNoSpace:
\begin{equation}
L_{P}(p_t) = \frac{1}{2}(log(p_t)-log(1-p_t))^2
\end{equation}
Figure \ref{fig:lossfunc} illustrates the behavior of the \lossFuncName  compared to the more common cross entropy loss. 
As can be observed, the \lossFuncName gives higher weights to examples where the probe classifier can predict the protected attribute well. The minimum occurs at $p_t=0.5$, where there is ambiguity for the probe classifier in determining the label. 
Eq. \ref{eq:AdvCE} is now modified as :
\begin{equation}
x_{p} = x - \lambda_P\cdot\nabla_{x}L_P(f(x), z) ,
\label{eq:projectiveEq}
\end{equation}
The gradient of the projective loss can be expressed as:
\begin{equation}
    \nabla_x L_{P} =  f(x)\nabla_x f^\top
\end{equation}

We show in Appendix \ref{sec:Theorem1proof} that Eq.\ \ref{eq:projectiveEq} with the \lossFuncName and a specific $\lambda_P = \frac{1}{\|\nabla_x f\|^2}$ yields a projection of the embedding vectors on the local linear model of each embedding.

\paragraph{Special Case of a Linear Probe Classifier.}
We now analyze the special case where $f$ is a linear classifier.
Given a linear classifier  $f(x) = x^\top \theta $ where $ \theta \in \mathbb{R}^d$  and a logistic function $\sigma(f) = \frac{1}{1+e^{-x^\top\theta}}$ to produce the probability $p_t$, we calculate the gradients of the \lossFuncName as:
\begin{equation}\label{eq:L_pr-chainrule}
\nabla_x L_{P} =  (x^\top\theta)\theta^\top
\end{equation}

Normalizing $\theta$ by setting $\lambda_P = \frac{1}{\|\nabla_x f\|^2} = \frac{1}{\theta^\top\theta}$ in Eq.~\ref{eq:projectiveEq} yields the orthogonal projection formula:
\begin{equation}
    x_{p} = x - (\frac{x^\top\theta}{\theta^\top\theta})\theta^\top
\label{eq:updateLp}
\end{equation}

This is also known as the null space projection which is used in INLP \cite{DBLP:conf/acl/RavfogelEGTG20}. INLP is a special case of our method when using a linear probe classifier. Unlike INLP, which obtains the projected embeddings by identifying the null space of a linear classifier, our method utilizes the gradients of neural network classifiers to obtain the projected embeddings.

INLP has been shown to be effective in removing sensitive information from neural representations \cite{DBLP:conf/acl/RavfogelEGTG20}. However, as highlighted by \citet{kumar2022probing}, a limitation of this approach is that each step of the projection operation decreases the norm of the representation, leading to its eventual reduction to zero as the number of steps increases. Our proposed method, \methodNameNoSpace, addresses this issue by utilizing a non-linear probe in the projection process, which does not reduce the rank of the representations. Thus, the removal of sensitive information is performed with minimal loss of other information as demonstrated in Section \ref{sec:eval_extrinsic}.\\

\subsection{\methodFullName}
\label{sec:implementation}

\algrenewcommand\algorithmicrequire{\textbf{Input:}}
\algrenewcommand\algorithmicensure{\textbf{Output:}}
\begin{algorithm}[t]
\caption{\methodFullName (\methodName)}
\label{alg:methodName}
\begin{algorithmic}
\Require{Model representations $X$, protected attribute $Z$, Stopping Criteria $S_c$}
\Ensure{New representations $X_{clean}$, probes list $F$}
\State $X_{0} \gets X$
\State $N \gets 0$
\State $F \gets [~]$
\While{(not $S_c$)}
\State $f \gets $TrainClassifier($X_N,Z$)
\State $F.append(f)$ 
\State $X_{N+1} \gets \{\}$
\For{$ x \in X_{N}$}
\State \texttt{{$x_{p} = x - \frac{\nabla_{x}L_P(f(x), z)}{\|\nabla_x f\|^2}$}}
\State $X_{N+1} \gets \{x_{p}\}\cup X_{N+1}$
\EndFor
\State $N \gets N+1$
\EndWhile\\
\Return $X_N , F$
\end{algorithmic}
\end{algorithm}
In this section we present our algorithm, \methodFullName (\methodNameNoSpace), for removing information of a discrete\footnote{This work primarily addresses the removal of discrete protected attributes (e.g., gender) information. However, in Appendix \ref{app:discrete} we show it can be adapted for continuous attributes (e.g., age).} attribute Z for a set of vectors X. Algorithm \ref{alg:methodName} presents the \methodName algorithm, which begins by training a classifier $f_1$ on the original representations X to predict a property Z. The projected representations $X^1_{p}$ are obtained by applying Eq.~\ref{eq:projectiveEq} to the original representations $X$. Since there are often multiple hypersurfaces that can capture sensitive attribute information, this process is repeated iteratively, each time using a newly trained classifier on the previous projected representations. The optimal number of iterations and the stopping criteria are determined with metrics such as accuracy or fairness. The relationship between the number of iterations and these metrics is explored in Section \ref{sec:num_it}.
\section{Experiments}
\label{sec:experiments}
In this section we compare competing methods for bias removal with the \methodName algorithm in both  intrinsic (Section \ref{sec:eval_intrinsic}) and extrinsic evaluations (Section \ref{sec:eval_extrinsic}), which are common in the literature on bias removal.

\subsection{Compared Methods}
We compare \methodName with several methods for bias mitigation, including a baseline (\textbf{Original}) without any concept-removal procedure.
\begin{description}

\item[\textbf{INLP}] \cite{DBLP:conf/acl/RavfogelEGTG20}, an iterative method that removes the protected information by projecting on the null space of linear classifiers.

\item[\textbf{RLACE}]  \cite{DBLP:conf/icml/RavfogelTGC22}, which removes linear concepts from the representation space as a constrained version of a minimax game where the adversary is limited to a fixed-rank orthogonal projection.

\item[\textbf{Kernelized Concept Erasure
(KCE)}]  \cite{ravfogel-etal-2022-adversarial}, which proposes a kernelization of a linear minimax game for concept erasure.
\end{description}

\subsection{Setup}
In each experiment, we utilize a a one-hidden layer neural network with ReLU activation as the attribute classifier for \methodName algorithm. Then we perform 5 runs of \methodName and competing methods with random initialization and report mean and standard deviations. 
Further details on implementation and hyperparameter tuning are provided in Appendix \ref{app:experiment}.
\subsection{Intrinsic Evaluation}
\label{sec:eval_intrinsic}

We begin by evaluating our debiasing method on GloVe \cite{pennington-etal-2014-glove} word embeddings, as it has been previously shown by \citet{bolukbasi2016man} that these embeddings contain unwanted gender biases. Our goal is to remove these biases. We replicate the experiment performed by \citet{DBLP:conf/acl-wnlp/GonenG19} and use the training and test data of \citet{DBLP:conf/acl/RavfogelEGTG20}, where the word vectors are labeled with their respective bias: male-biased or female-biased. See
Appendix \ref{app:experiment} for more details on the experimental setting.\\ 

\subsubsection{Embeddings Classification} 
\label{sec:emb_dist}
\begin{table}[t]
\centering
\setlength{\tabcolsep}{10pt}
\renewcommand{\arraystretch}{1.2}
\begin{tabular}{lcc}

\toprule
        
 Method & \shortstack{Leakage\\ Linear}$\downarrow$ & \shortstack{Leakage\\ Non-Linear}$\downarrow$ \\

\midrule
Original &  100$\pm{\scriptstyle0.00}$      &  100$\pm{\scriptstyle0.00}$   \\

INLP     &      55.03$\pm{\scriptstyle1.29}$  &  94.42$\pm{\scriptstyle1.85}$   \\

RLACE    &      53.80$\pm{\scriptstyle1.37}$  &  92.53$\pm{\scriptstyle1.87}$   \\

KCE     &     60.01$\pm{\scriptstyle0.03}$    &  96.20$\pm{\scriptstyle1.30}$   \\

\methodName &  56.56$\pm{\scriptstyle4.25}$   & 69.89$\pm{\scriptstyle2.81}$   \\
\bottomrule
\end{tabular}
\caption{Gender leakage from  GloVe word embeddings using linear and non-linear classifiers.}
\label{table:emb_dis}
\end{table}

After applying the debiasing methods, we follow the evaluation approach proposed by \citet{DBLP:conf/acl-wnlp/GonenG19} and train new classifiers, a linear SVM and a non-linear SVM with RBF kernel, to predict gender from the new representations. We define \emph{leakage} as the accuracy of these classifiers. The results are shown in Table \ref{table:emb_dis}. As we can see, all methods are effective at removing linearly encoded information, as the leakage is very low. However, when using non-linear classifiers, all competing methods fail to eliminate leakage, including KCE.\footnote{\citet{ravfogel-etal-2022-adversarial} also demonstrated that KCE's attribute protection fails against other type of adversaries, even with the same kernel but different parameters.} Even though the adversary classifier used to calculate leakage (SVM-RBF) is different from the ReLU MLP employed in \methodNameNoSpace, our method is still the most effective at removing non-linearly encoded information. The results demonstrate the advantage of \methodName in eliminating non-linear information in word embeddings over competing methods.

\subsubsection{WEAT Analysis} 
\label{sec:weat}

\setlength{\tabcolsep}{4pt}
\renewcommand{\arraystretch}{1.2}
\begin{table}[ht]
\centering
\begin{tabular}{c c cc}
\toprule
 & Method & WEAT's d$\downarrow$& WEAT's p$\uparrow$\\
\midrule
\multirow{5}{*}{\rotatebox{90}{Math-art}} & Original &  $1.57\pm{\scriptstyle0.00}$ & $0.000\pm{\scriptstyle0.00}$ \\
                          &   INLP     &  $1.10\pm{\scriptstyle0.10}$ & $0.016\pm{\scriptstyle0.00}$ \\
                          &   RLACE    &  $0.80\pm{\scriptstyle0.01}$ & $0.062\pm{\scriptstyle0.00}$ \\
                          &   KCE     & $0.78\pm{\scriptstyle0.01}$ & $0.067\pm{\scriptstyle0.00}$ \\
                          &   \methodName     & $\textbf{0.73}\pm{\scriptstyle0.01}$ & $\textbf{0.091}\pm{\scriptstyle0.00}$ \\

\midrule

\multirow{5}{*}{\rotatebox{90}{Science-art}} & Original &  1.63$\pm{\scriptstyle0.00}$  & 0.000$\pm{\scriptstyle0.00}$\\
& INLP     &  1.08$\pm{\scriptstyle0.00}$  & 0.011$\pm{\scriptstyle0.00}$\\
& RLACE    &  0.77$\pm{\scriptstyle0.01}$ & 0.073$\pm{\scriptstyle0.003}$\\
& KCE     & 0.74$\pm{\scriptstyle0.00} $& 0.08$\pm{\scriptstyle0.00} $\\
& \methodName     & $\textbf{0.19}\pm{\scriptstyle0.01} $& $\textbf{0.64}\pm{\scriptstyle0.01}$ \\
\midrule
\multirow{5}{*}{\rotatebox{90}{Prof-family}} & Original &  1.69$\pm{\scriptstyle0.00}$ & 0.000$\pm{\scriptstyle0.00}$ \\
& INLP     &  1.15$\pm{\scriptstyle0.07} $& 0.007$\pm{\scriptstyle0.00}$ \\
& RLACE    &  0.78$\pm{\scriptstyle0.01}$ & 0.072$\pm{\scriptstyle0.00}$\\
& KCE     & 0.73$\pm{\scriptstyle0.01}$  & 0.090$\pm{\scriptstyle0.05}$ \\
& \methodName     & $\textbf{0.21}\pm{\scriptstyle0.00}$ & $\textbf{0.330}\pm{\scriptstyle0.00}$ \\
\bottomrule
\end{tabular}
\caption{WEAT test results.}
\label{table:weat}
\end{table}

The Word Embeddings Association Test \cite{caliskan2017semantics} is a measure of bias in static word embeddings, which compares the association of male and female related words with stereotypically male or female professions. 
We follow \citet{DBLP:conf/acl-wnlp/GonenG19} in defining the groups of male and females associated words. We represent the gender groups with three categories (1) art and mathematics; (2) art and science; and (3) career and family. We present the results of the WEAT test in Table \ref{table:weat}, including the d-value and the p-value (refer to \citet{caliskan2017semantics} for further information). 
We found that \methodName has the most effective debiasing effect on word embeddings compared to other methods.
\subsubsection{Semantic Similarity Analysis}
\label{sec:semantic}
In addition to mitigating bias in word embeddings, it is important to examine if any semantic content was damaged. We perform a semantic evaluation of the debiased word embeddings using  SimLex999 \cite{hill2015simlex}, an annotated
dataset of word pairs with human similarity
scores for each pair.
 As displayed in Table \ref{table:semantic1}, \methodName and other methods yield only a slight reduction in correlation.
To qualitatively assess the impact of IGBP on semantic similarity in GloVe word embeddings, we provide a random sample of words and their nearest neighbors before and after debiasing in Appendix ~\ref{app:word_emb_exp}. We observe minimal change to the nearest neighbors.

\subsection{Extrinsic Evaluation}
\label{sec:eval_extrinsic}
In this section we focus on evaluating \methodName in the context of classification tasks. We focus on tasks where we want to eliminate a concept from the representations to prevent the main classifier from using it, thus ensuring fair classification.
\setlength{\tabcolsep}{10pt}
\renewcommand{\arraystretch}{1.3}
\begin{table}[t]
\centering

\begin{tabular}{l c }
\toprule
 Method &  $\rho \uparrow$\\

\midrule
Original &  $0.400\pm{\scriptstyle0.000}$  \\
\midrule
INLP     &      $0.389\pm{\scriptstyle0.001}$  \\

RLACE    &      $0.389\pm{\scriptstyle0.001}$  \\

KCE     &     $0.393\pm{\scriptstyle0.001}$    \\

\methodName &  $0.387\pm{\scriptstyle0.001}$   \\
\bottomrule 
\end{tabular} 
\caption{Evaluation of semantic content using Simlex-999 dataset. The scores shown are the Pearson correlation coefficient between the similarity scores assigned by humans and those computed using the embeddings.}
\label{table:semantic1}
\end{table}
\subsubsection{Evaluation Metrics}

To measure extrinsic bias, we calculate the True Positive Rate Gap \emph{(TPR GAP)} to measure the differences in performance between the different protected attribute groups. 

\begin{table}[b]
\setlength{\tabcolsep}{5pt}
\centering
\begin{tabular}{l  c  c}
\toprule
  & DIAL & BIOS\\
\midrule
Main Task & Sentiment & Profession\\
Attribute & Race & Gender\\
Size  & 100K/ 8K/ 8K & 255K/ 39K/ 43K\\
\bottomrule 
\end{tabular} 
\caption{Dataset characteristics. 
Main classification task, protected attribute, and sizes of training, development, and test sets, in each dataset. }
\label{table:datasets}
\end{table}
\[\mathrm{TPR_{z,y} = P(\hat{Y}=y|Z=z,Y=y)}\] \[\mathrm{GAP_{TPR}^{z,y} = TPR_{z,y} - TPR_{z',y}}\]
To assign a single bias measure across all values of $y$, we follow \citet{romanov-etal-2019-whats} and calculate the root mean square \(\mathrm{GAP_{TPR}^{z}}\) in order to obtain a single bias score over all labels $y$:  
\begin{equation}
    \mathrm{GAP_{TPR}^{z} = \sqrt{\frac{1}{|N|}\sum_{y\in N}(GAP_{TPR}^{z,y})^2}}
    \label{eq:tpr_gap}
\end{equation}

For example, in a sentiment analysis task, it is important for the model to have equal performance across all demographic groups, as measured by the TPR. This ensures that the model's predictions are fair and not biased towards any particular group. 

We report two common metrics for measuring bias in representations: : (1) \emph{Leakage}, as described in Section \ref{sec:eval_intrinsic}; (2) \emph{Minimum Description Length (MDL) Compression} \cite{voita-titov-2020-information}, which serves as an indicator of the extent to which certain biases can be extracted from a model's representations \cite{orgad-belinkov-2022-choose}. A higher compression score indicates that it is easier to extract the protected attribute from the model's representation. \citet{orgad-etal-2022-gender} found that this metric highly correlates with extrinsic bias metrics. We use a ReLU MLP of two-hidden layers of size 512 as the probe classifier.
 We provide more details about these metrics in Appendix~\ref{app:extrinsic_eval}. 

\renewcommand{\arraystretch}{1.3}
\setlength{\tabcolsep}{5pt}
\begin{table*}[t]

\begin{subtable}{\textwidth}
\adjustbox{width=\textwidth}{%
\begin{tabular}{p{2cm}cccc|cccc}
\toprule
{} &  \multicolumn{4}{c}{BERT} & \multicolumn{4}{c}{RoBERTa} \\
\cline{2-5} \cline{6-9}
Method & Acc $\uparrow$ & GAP$_{\small{TPR}} \downarrow$ & Leakage$ \downarrow$ & C $\downarrow$ & Acc $\uparrow$ & GAP$_{\small{TPR}}\downarrow$ & Leakage $\downarrow$ & C $\downarrow$ \\
\midrule
Original &  79.89$\pm{\scriptstyle0.06}$       &  15.55$\pm{\scriptstyle0.16}$          & 99.32$\pm{\scriptstyle0.11}$ & 30.81$\pm{\scriptstyle0.18}$ &  79.08$\pm{\scriptstyle0.05}$       &  19.26$\pm{\scriptstyle0.40}$          & 97.25$\pm{\scriptstyle0.11}$ & 11.09$\pm{\scriptstyle0.00}$ \\
\midrule
INLP    &      75.65$\pm{\scriptstyle0.03}$   &  13.52$\pm{\scriptstyle0.13}$        &  95.77$\pm{\scriptstyle1.42}$ & 7.76$\pm{\scriptstyle0.60}$  &      76.75$\pm{\scriptstyle0.05}$   &  10.71$\pm{\scriptstyle0.05}$        &  81.29$\pm{\scriptstyle1.04}$ & 1.78$\pm{\scriptstyle0.03}$\\
RLACE   &      79.77$\pm{\scriptstyle0.07}$   &  13.54$\pm{\scriptstyle0.13}$        & 98.55$\pm{\scriptstyle0.19}$   & 13.31$\pm{\scriptstyle0.99}$  &      78.57$\pm{\scriptstyle0.07}$   &  11.82$\pm{\scriptstyle0.27}$        & 90.87$\pm{\scriptstyle1.90}$   & 2.80$\pm{\scriptstyle0.19}$ \\
KCE     &     78.16$\pm{\scriptstyle0.05}$    &  13.65$\pm{\scriptstyle0.12}$        & 97.35$\pm{\scriptstyle0.15}$   & 11.67$\pm{\scriptstyle1.01}$ & 78.54$\pm{\scriptstyle0.04}$    &  13.94$\pm{\scriptstyle0.18}$        & 96.60$\pm{\scriptstyle0.21}$       & 6.57$\pm{\scriptstyle0.84}$\\
\methodName &  78.80$\pm{\scriptstyle0.19}$ & \textbf{9.87$\pm{\scriptstyle0.25}$} & \textbf{69.72$\pm{\scriptstyle2.56}$}  & \textbf{1.66$\pm{\scriptstyle0.08}$} & 77.49$\pm{\scriptstyle0.04}$     & \textbf{9.45$\pm{\scriptstyle0.04}$}& \textbf{65.71$\pm{\scriptstyle0.44}$} &\textbf{1.54$\pm{\scriptstyle0.01}$} \\
\bottomrule
\end{tabular}
}
\caption{Frozen models}
\label{table:BiasInBios_a}
\end{subtable}

\medskip

\begin{subtable}{\textwidth}
\adjustbox{width=\textwidth}{%
\begin{tabular}{p{2cm}cccc|cccc}
\toprule
{} &  \multicolumn{4}{c}{BERT} & \multicolumn{4}{c}{
 RoBERTa} \\
\cline{2-5} \cline{6-9}
Method & Acc $\uparrow$ & GAP$_{\small{TPR}} \downarrow$ & Leakage$ \downarrow$ & C $\downarrow$ & Acc $\uparrow$ & GAP$_{\small{TPR}}\downarrow$ & Leakage $\downarrow$ & C $\downarrow$ \\
\midrule

Original &  85.15$\pm{\scriptstyle0.04}$       &  13.45$\pm{\scriptstyle0.11}$          & 98.49$\pm{\scriptstyle0.02}$ & 13.58$\pm{\scriptstyle0.08}$ &  
84.09$\pm{\scriptstyle0.10}$ &  14.57$\pm{\scriptstyle0.16}$ & 99.02$\pm{\scriptstyle0.01}$ & 17.28$\pm{\scriptstyle0.00}$ \\
\midrule

INLP    &      85.08$\pm{\scriptstyle0.03}$   &  12.71$\pm{\scriptstyle0.04}$        &  97.08$\pm{\scriptstyle0.00}$ & 6.01$\pm{\scriptstyle0.00}$  &      83.78$\pm{\scriptstyle0.05}$ &  14.18$\pm{\scriptstyle0.10}$ &  97.74$\pm{\scriptstyle0.80}$ & 10.42$\pm{\scriptstyle0.01}$\\

RLACE   & 85.12$\pm{\scriptstyle0.04}$   &  12.93$\pm{\scriptstyle0.14}$  & 98.26$\pm{\scriptstyle0.05}$  & 8.87$\pm{\scriptstyle0.01}$   &      83.85$\pm{\scriptstyle0.10}$  &  14.21$\pm{\scriptstyle0.05}$  & 98.84$\pm{\scriptstyle0.02}$  & 11.31$\pm{\scriptstyle0.01}$ \\

KCE     &     84.86$\pm{\scriptstyle0.03}$    &  12.81$\pm{\scriptstyle0.12}$        & 98.70$\pm{\scriptstyle0.04}$   & 9.43$\pm{\scriptstyle0.01}$ & 83.94$\pm{\scriptstyle0.04}$    &  14.30$\pm{\scriptstyle0.08}$        & 98.33$\pm{\scriptstyle0.02}$       & 13.04$\pm{\scriptstyle0.02}$\\

\methodName &  83.70$\pm{\scriptstyle0.05}$ & \textbf{9.63$\pm{\scriptstyle0.18}$} & \textbf{65.47$\pm{\scriptstyle0.40}$} &\textbf{1.54$\pm{\scriptstyle0.01}$} & 82.88$\pm{\scriptstyle0.13}$     & \textbf{10.78$\pm{\scriptstyle0.10}$}& \textbf{65.73$\pm{\scriptstyle0.40}$} &\textbf{1.53$\pm{\scriptstyle0.01}$} \\

\bottomrule
\end{tabular}
}

\caption{Finetuned models}
\label{table:BiasInBios_b}
\end{subtable}

\medskip

\caption{Evaluation results on the test set in Bias in Bios dataset with BERT and RoBERTa as encoders.  C is the compression of the probing classifier. The best result is highlighted with bold if the difference over the next-best method is statistically significant (based on T-test; $p < 0.05$).}
\label{table:BiasInBios}
\end{table*}

\subsubsection{Datasets}
We experiment with the following two datasets  (Table \ref{table:datasets} provides a brief summary):

\paragraph{Bios.}
\label{sec:biasinbios}
The Bias in Bios dataset \cite{10.1145/3287560.3287572} contains 394K biographies. The task is to predict a person’s occupation (out of 28 professions) based on their biography. Gender annotations are provided for each biography, and we aim to eliminate any gender-related information encoded in the representations. We split to training, development, and test sets following \citet{10.1145/3287560.3287572}.
The pre-trained BERT model \cite{DBLP:conf/naacl/DevlinCLT19} is used as the encoder and the final hidden layer's [CLS] token is used as a representation for the biography. To ensure that the results are not model-specific, the experiment is replicated using the pre-trained RoBERTa model \cite{DBLP:journals/corr/abs-1907-11692} as the encoder. Additionally, the experiment is conducted with fine-tuned models.

\paragraph{DIAL.}
\label{sec:DIAL}

Dialectal tweets (DIAL) is a corpus of tweets collected by \citet{blodgett-etal-2016-demographic}, where the task is to predict the sentiment of the tweet (positive or negative). Each tweet is associated with the sociolect of the author (African American English or Standard American English), which is a proxy for the racial identity of the author. Following \citet{DBLP:conf/acl/RavfogelEGTG20} setup, we filter the corpus and split the data into training, development, and test sets. We use The DeepMoji model \cite{felbo2017} as an encoder to produce representations.

\subsubsection{Results}
\label{sec:results}

\paragraph{Bios.} 
The results from the Bias in Bios experiment are summarized in Table \ref{table:BiasInBios}. With both BERT and RoBERTa frozen pre-trained models (Table \ref{table:BiasInBios_a}), it can be observed that while INLP reduces TPR-GAP, it degrades overall performance in the process. This may be due to INLP's limitation of decreasing representation's rank each step. RLACE and KCE lead to a reduction in the TPR-GAP but the value remains elevated.
On the other hand, our proposed method \methodName significantly reduces TPR-GAP while only causing a slight decrease in main task accuracy. Furthermore, in terms of intrinsic bias, \methodName is distinguished by its effectiveness at decreasing non-linear leakage and compression. As for the results with fine-tuned models (Table \ref{table:BiasInBios_b}), it shows similar results. Other competing methods only exhibit minimal reduction in TPR-GAP, whereas our approach, \methodNameNoSpace, succeeds in enhancing fairness and eliminating leakage.

\paragraph{DIAL.} 
Table \ref{table:DIAL} presents a summary of the results obtained on DIAL dataset. The results show that applying \methodName leads to a significant reduction in the TPR-GAP, with a statistically significant difference compared to the other methods, while maintaining a level of accuracy comparable to the original model. In terms of intrinsic evaluation of the representations, both INLP and RLACE, reduce leakage and compression
, but not to the same extent as \methodNameNoSpace. While KCE fails to reduce bias.

On a whole, we found that our proposed method outperforms competing methods empirically in terms of reducing extrinsic and intrinsic bias, and offers a more balanced accuracy--fairness tradeoff.

\renewcommand{\arraystretch}{1.5}
\setlength{\tabcolsep}{5pt}
\begin{table}[t]
\resizebox{\columnwidth}{!}{%
\begin{tabular}{p{1.5cm}cccc}
\toprule

Method  &   Accuracy$\uparrow$  & GAP$_{\small{TPR}}\downarrow$ &  Leakage$\downarrow$ &  $C\downarrow$ \\
\midrule
Original &  73.89$\pm{\scriptstyle0.04}$   &  30.19$\pm{\scriptstyle0.02}$   & 75.67$\pm{\scriptstyle0.11}$    & 2.14$\pm{\scriptstyle0.00}$ \\
\hline
INLP    &      69.59$\pm{\scriptstyle1.14}$ &  17.59$\pm{\scriptstyle0.77}$  &  62.28$\pm{\scriptstyle1.24} $  & 1.63$\pm{\scriptstyle0.00}$\\
RLACE   &     72.98$\pm{\scriptstyle0.34}$ &  13.53$\pm{\scriptstyle1.89}$   & 61.92$\pm{\scriptstyle1.67}$   & 1.62$\pm{\scriptstyle0.00}$ \\
KCE     &     72.92$\pm{\scriptstyle0.24}$  &  29.25$\pm{\scriptstyle0.81}$  & 73.63$\pm{\scriptstyle1.66}$   &  2.12$\pm{\scriptstyle0.00}$\\
\methodName &  72.87$\pm{\scriptstyle0.31}$  & \textbf{9.23$\pm{\scriptstyle0.04}$} & \textbf{56.53$\pm{\scriptstyle3.57}$} & \textbf{1.43$\pm{\scriptstyle0.00}$}   \\
\bottomrule
\end{tabular}
}
\caption{Evaluation results on the test set of DIAL. The notation used here is consistent with Table \ref{table:BiasInBios}.}
\label{table:DIAL}
\end{table}
\section{Analysis}
We conduct a series of analyses of our proposed method: an examination of probe's complexity impact on debiasing and an analysis of the effect of number of iterations on performance.

\subsection{Effect of Probe Complexity}
\label{sec:probe_complexity}
\begin{figure}[htb]
\centering

   \includegraphics[width=1\linewidth]{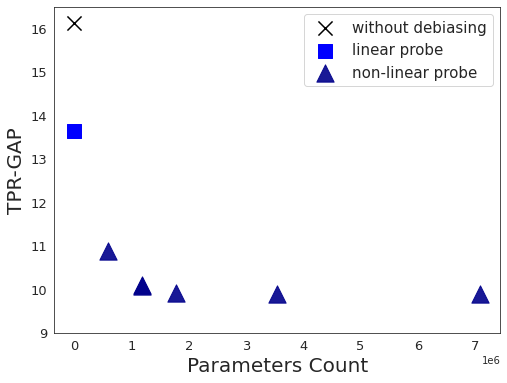}

\caption{The TPR-GAP results obtained by applying \methodName with different probe classifier architectures on the Bias in Bios dataset.}
\label{fig:nonlinear_bios}
\end{figure}

\methodName proved to be superior to linear information-removal methods in the experiments presented in Section \ref{sec:experiments}. To further investigate the potential of reducing bias, we will explore the use of more complex non-linear probes by varying the width and depth of the neural network used as a probe in \methodName\footnote{For further details on the probes architecture used, see App.\ref{app:benefits_nonlin}}. Figure \ref{fig:nonlinear_bios} shows the TPR-GAP score after applying 50 iterations of \methodName on Bias in Bios dataset\footnote{ In Appendix \ref{app:analysis} we show similar results on DIAL dataset.}. As we can see, there is a noticeable reduction in TPR-GAP when using non-linear probes instead of linear probe. 
Applying \methodName with a growing complexity of probe classifiers (moving from left to right) also result in a lower TPR-GAP. However, the reduction is not significant. 
We also report that the more complex the probe, the greater the accuracy drop, but not at a significant value: The maximum accuracy drop was 1.20\%. To conclude, based on the results in Section \ref{sec:experiments} and this expirement, one-hidden layer probe is enough to reduce bias related to gender and race in text representation. Using more complex probes may offer some additional benefits, but the improvement will be limited.

\begin{figure}[b]
\includegraphics[width=0.48\textwidth,height=0.25\textwidth]{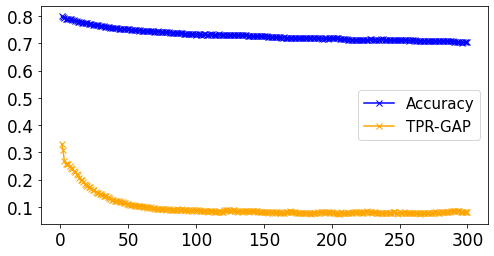}
\vspace{-0.5cm}
\caption{ Mean accuracy and TPR-GAP results versus number of iterations on DIAL dataset, averaged over 5 different random seeds. }
\label{fig:iterations}
\end{figure}
\subsection{Perforamnce -- Fairness Tradeoff}
\label{sec:num_it}
One of the key factors that influence the effectiveness of our method is the number of iterations used. Varying the number of iterations and measuring the resulting changes in the TPR-GAP and downstream task accuracy on the DIAL development set shows that the number of iterations had a significant impact on attribute removal in the early stages (Figure \ref{fig:iterations}), but eventually reached a plateau. Increasing the number of iterations also harmed downstream task accuracy, but the decrease was gradual. A similar experiment on Bias in Bios (Appendix \ref{app:analysis}) showed the same trend. The results suggest that the balance between performance and fairness can be controlled by adjusting the number of iterations or by implementing appropriate stopping criteria.

\section{Conclusion}
We presented a gradient-based method for the erasure of non-linearly encoded concepts in text representations. Its ability to remove non-linear information makes it particularly useful for addressing the complex biases that may be present in text representations learned through complex models. We empirically show the effectiveness of our approach to mitigate social biases in representations, thereby improving fairness in models' decision-making. 

Beyond mitigating bias, the \methodFullName method has the potential to be applied in a wide range of other contexts, such as increasing model interpretability by applying causal interventions, adapting models to new domains by removing domain-specific information and ensuring privacy by removing sensitive information. In future work, we plan to explore these and other potential applications of the proposed method.

\section*{Limitations}
The proposed method has limitations in its dependence on the accuracy and performance of the probe classifier as noted in \cite{belinkov-2022-probing}, and may be limited in scenarios where the dataset is small or lacks sufficient information about the protected attribute. Additionally, this approach increases inference time due to the use of a sequential debiasing classifiers. In future work, we aim to find a single probe that eliminates non-linear leakage. Finally, the proposed method aims to eliminate information about a protected attribute in neural representations. While it may align with fairness metrics such as demographic parity, it is not specifically designed to ensure them.

\section*{Ethical Considerations}
 Ethical considerations are of utmost importance in this work. It is essential to exercise caution and consider the ethical implications when using this method, as it has the potential to be applied in situations where fair and unbiased decision-making is critical. It is important to thoroughly evaluate the effectiveness of the method in the specific context in which it will be used, and to carefully consider the data, fairness metrics, and overall application before deploying it. It is worth noting that our method is limited by the fact that gender is a non-binary concept and that it does not address all forms of bias, and further research is necessary to identify and address these biases. Additionally, it is important to consider the potential risk of inadvertently increasing bias through reversing the direction of the debiasing operation in the algorithm. It is crucial to be mindful of the potential impact of this method and to approach its use with caution and care.
\section*{Acknowledgment}
This project was supported by an AI Alignment grant from Open Philanthropy, the Israel Science Foundation (grant No.\ 448/20), and an Azrieli Foundation Early Career Faculty Fellowship.
\bibliography{anthology,custom}
\bibliographystyle{acl_natbib}
\cleardoublepage
\section*{Appendix}
\appendix
\section{Continuous Attributes}
\label{app:discrete}
While this work focuses on discrete attribute information-removal, we explain briefly how it can be adapted for regression problems, where the attribute is continuous (e.g., age). In discrete attribute classification tasks, Given that $f$ is the classifier, \methodName is designed to transform each vector $x$ to $x'$ onto the decision boundary of $f$ such that $f(x')=0$. In the continuous case, where $f$ is the attribute regressor, \methodName aims to achieve a similar result, with the goal of projecting each vector $x$ onto a point $x'$ such that $f(x')=0$. Hence, each input, x, is regressed to a non-informative value of zero, meaning that the input is stripped of its information content.


\section{Reversal Gradient of Cross Entropy}
\label{app:math_grad_CE}
Let us consider a non-linear model $f(x)$ followed by a logistic function to obtain the probabilty $p = \frac{1}{1+e^{f(x)}}$.
Then the gradient of  $L_{CE}(p_t)$ when $y = 1$ is :
\begin{multline}
        \nabla_x{L_{CE}} = \frac{\partial L_{CE}}{\partial p_t} \frac{\partial p_t}{\partial f}\frac{\partial f}{\partial x} \\
        \hspace{2.7em}    = \frac{-1}{p_t} \: p_t(1-p_t)  \: \nabla_x{f}\\
         = -(1-p_t) \:\nabla_x{f}\\
\label{eq:chain-rule}
\end{multline}

and $\nabla_x{L_{CE}} = +(1-p_t) \:\nabla_x{f}$ when $y = -1$.
\section{Local Linear Model Projection}
\label{sec:Theorem1proof}

We will now show how the projective loss update step projects each sample to its local linear model boundary. This will facilitate the probe being oblivious to the protected attribute.

\paragraph{Local Linearity.} First, we will show that a trained ReLU neural net probe divides the embedding space into sub-regions, where in each sub-region it behaves as a linear model. We will demonstrate that we can obtain the local linear model for each embedding.
Let us consider a non-linear probe composed of one-hidden layer with ReLU as an activation function:\footnote{The extension for multiple hidden layers and different piece-wise linear activation functions is straightforward.}
\begin{multline}
        z = x^\top W, \\
        h = ReLU(z), \\
         f = h^\top\theta, \\
        p = \frac{1}{1+e^{-f(x)}}\\
\end{multline}
The activation function ReLU acts as an element-wise scalar (0 or 1) multiplication, hence h can be written as:
\begin{equation}
    h = a \odot z
\end{equation}
where $a$ is a vector with (0,1) entries indicating the slopes of ReLU in the corresponding linear regions where z fall into.
Let us define a diagonal matrix $D$:
\begin{equation}
    D = diag(a)
\end{equation}
Then,
\begin{equation}
    h = Dz
\end{equation}
since doing element wise multiplication with a vector $a$ is the same as multiplication by the diagonal matrix $D$. It is now possible to express the output in each sub-region in matrix form as follows:
\begin{multline}
         f = h^\top\theta \\
      = (DWx)^\top\theta  \\
      = x^\top (DW)^\top\theta \\ 
      \label{eq:f=x..}
\end{multline}
   
$D$ expresses the ReLU function, so naturally it depends on $Wx$, but since the weights of the probe are frozen/constant and we are doing the calculation for the sub-region where the slope of the ReLU function is constant, we can assume that $D$ is not dependent on $x$ in this sub-region.Thus, in each sub-region $r$ defined by the classifier, the local linear model for this sub-region is $\theta_r$ defined bellow :

\begin{equation}
    \theta_r =  (DW)^\top\theta \\
\end{equation}
We can obtain the vector $\theta_r$ with the gradient of $f$:
\begin{equation}
    \nabla_x{f} = (DW)^\top\theta \\
\label{eq:dfdx}
\end{equation}

Applying the chain rule with $L_{P}(p_t)$ as in Eq.~\ref{eq:chain-rule} for each sub-region:
\begin{multline}
\nabla_x{L_{P}}= \frac{\partial L_{P}}{\partial p_t} \frac{\partial p_t}{\partial p} \frac{\partial p}{\partial f}\frac{\partial f}{\partial x} \\ 
= \frac{log(\frac{p_t}{1-p_t})}{p_t(1-p_t)}(-1)^yp(1-p)\nabla_x{f}^\top   \\ 
\hspace{-2.2em} = (-1)^{-y}f(x)(-1)^{y}\nabla_x{f}^\top \\
\hspace{-2.2em} = f(x)\nabla_x{f}^\top \quad \\
\hspace{1em} =  x^\top (DW)^\top\theta((DW)^\top\theta)^\top \text{(Using Eq.~\ref{eq:f=x..},\ref{eq:dfdx})}\\
\hspace{-3.8em}= (x^\top\theta_r)\theta_r^\top \\
\end{multline}

Again, we can obtain $\theta_r$ from the gradient and divide the term with $\frac{1}{\theta_r^\top\theta_r}$ to get the linear projection of each sub-region to its linear model null space
\begin{equation}
    x_{p} = x -  (\frac{x^\top\theta_r}{\theta_r^\top\theta_r})\theta_r^\top
\end{equation}

\section{Experiment}
\label{app:experiment}
This section provides additional details on the experimental setup and results.

\subsection{Implementaion details}
\paragraph{\methodName stopping criteria.} In order to balance the trade-off between reducing extrinsic and intrinsic bias while preserving accuracy (as can be seen in Section \ref{sec:num_it}), we have established a stopping criterion for our proposed method, \methodName. The criterion is based on two factors: the accuracy of a newly trained probe classifier on the protected attribute, and the main task accuracy on the development set. Specifically, we run Algorithm \ref{alg:methodName} until the newly trained probe classifier acheives within 2\% above-majority accuracy, or until the main task accuracy on the development set drops below a threshold of 0.98 of the original main task accuracy. Through empirical analysis on the development set, we have determined that this threshold yields good results for all extrinsic evaluation experiments. However, it is worth noting that this stopping criterion may be adjusted based on specific requirements for each case.

\paragraph{\methodName classifier type.}For all experiments, we use a a ReLU MLP as the attribute classifier with a single-hidden layer of the same size as the input dimension. We train the classifier with AdamW optimizer \cite{loshchilov2018decoupled} with learning rate of $2e^{-4}$ and batch size of 256.\\
 Applying the algorithm for training on DIAL dataset takes about 0.5-1 hour and 1-3 hours on Bias in Bios on NVIDIA GeForce RTX 2080 Ti.

\paragraph{Compeing methods implementation and hyperparameters.} For competing methods,  we follow their implementations that can be found here\footnote{\url{https://github.com/shauli-ravfogel/nullspace_projection}\\\url{https://github.com/shauli-ravfogel/rlace-icml}\\\url{https://github.com/shauli-ravfogel/adv-kernel-removal}}. We run the algorithms until the specific type of leakage they were trying to eliminate was no longer present. For KCE we choose RBF kernel following their selection in their paper for Bias in Bios task. We tried multiple kernels but found that RBF yeilds better results. The results of RLACE are different than those in the original paper because they used only the first 100K of training samples and applied a PCA transformation to reduce dimensions down to 300 due to the high computation time. However, we wanted to make fair comparison so we did not reduce the size of training set or the dimensionality.

\paragraph{Models.} We used the pre-trained BERT and RoBERTa base models by Huggingface that have 110M and 123M parameters. They were fine-tuned on the proffesion prediction task in Bias in Bios using a stochastic gradient descent (SGD) optimizer with a learning rate of $5e^{-4}$, weight decay of $1e^{-6}$, and momentum of $0.90$. We trained for $30,000$ batches of size $10$.

\subsection{GloVe Word Embeddings Experiment}
\label{app:word_emb_exp}
We provide details about the experimental settings in the static word vectors experiment \ref{sec:eval_intrinsic}. We follow \citet{DBLP:conf/acl/RavfogelEGTG20} and use uncased GloVe word embeddings of 150,000 most common words. We project all vectors on \(\Vec{he} - \Vec{she}\) direction, and select the 7500 most male-biased and female biased words. Using the same training–development–test split as \citet{DBLP:conf/acl/RavfogelEGTG20}, we subtract the gender-neutral words and end up with a training set of 7350, an evaluation set of 3150, and a test set of 4500.

\subsubsection{Additional intrinsic evaluation}
\label{app:additional_eval}
We evaluate bias-by-neighbors which was proposed by \cite{DBLP:conf/acl-wnlp/GonenG19} and the list of professions from \cite{bolukbasi2016man}. We determine the correlation between bias-by-projection and bias-by-neighbors by calculating the percentage of top 100 neighboring words for each profession that were originally biased-by-projection towards a specific gender. Our results show mean correlation of 0.598, which is lower than the previous correlation of 0.852. In comparison, after applying INLP we find a correlation of 0.73. This suggests that while some bias-by-neighbors still remains, the debiasing effect of \methodName is significant.

\subsubsection{Nearest Neighbors}
We demonstrated in Section \ref{sec:semantic} that debiasing using \methodName did not cause significant harm to the GloVe word embedding space as per the SimLex999 test results. To further support this, in Table \ref{table:nearest}, we present the closest neighbors to 10 randomly sampled words from the vocabulary, both before and after our debiasing procedure, as a qualitative illustration.
\label{app:nearest}

\begin{table}[H]
\resizebox{\linewidth}{!}{%
  \begin{tabular}{|l|l|l|}
  
    \hline
    
    Word & Neighbors Before & Neighbors After \\
    \hline
    period & periods,during,time & periods,during,time \\
    actual & exact,any,same & exact,any,same \\
    markers & marker,marking,pens & marker,marking,pens \\
    photoshop & adobe,illustrator,indesign  & adobe,illustrator,indesign \\
    commands & command,execute,instructions & command,execute,scripts \\
    adapted & adaptation,adapting,adapt & adaptation,adapting,adapt \\
    called & known,which,that & known,which,also \\
    vital & crucial,important,essential & crucial,important,essential\\
    heritage & cultural,historic,historical & cultural, historic,historical\\
    mood & moods,feeling,feel & moods,feeling,feel\\
    \hline
  \end{tabular}
  }
    \caption{3-neighbors of random words in GloVe embedding space before and after \methodName debiasing.}
\label{table:nearest}

\end{table}

\subsection{Extrinsic Evaluation Experiments}
\label{app:extrinsic_eval}
\subsubsection{Metrics}
We provide additional details on the metrics used in Section \ref{sec:eval_extrinsic}
\label{app:metrics}
\paragraph{Main task model.} We use sklearn's SVM \cite{scikit-learn} for the main task predictions on DIAL experiment, and sklearn's logistic regression for Bias in Bios which is a multi-label classification task.
\paragraph{Leakage and MDL Compression.} MDL is an
information-theoretic probing which measures how efficiently a model can extract information about the labels from the inputs . In this work, we employ the online coding approach  \cite{voita-titov-2020-information} to calculate MDL. We estimate MDL following \citeauthor{voita-titov-2020-information}'s online coding $L_{online}$ and calcuate the \textbf{compression}, \textbf{C}, which is compared against uniform encoding $L_{uniform}$ which does not require any learning from data. \[C = \frac{L_{uniform}}{L_{online}}\]
We evaluate our models using an online code probe, which is trained on fractions of the training dataset: [2.0, 3.0, 4.4, 6.5, 9.5, 14.0, 21.0, 31.0, 45.7, 67.6, 100]. Then we calculate leakage as the probe's accuracy on test set when trained on the entire training set.
We use a MLP with two-hidden layer of size 512 and ReLU activation as the probe classifier. This decision was made to stay consistent with previous work which employed MDL \cite{mendelson-belinkov-2021-debiasing} and to have a different and more powerful adversary than the one used in \methodName.
\section{Analysis}

\label{app:analysis}
\subsection{Biographies representation}
We present the t-SNE \cite{van2008visualizing} projections of the biographies representations of BERT before and after applying IGBP.

\begin{figure}[htb]
\centering
\includegraphics[width=0.48\textwidth,height=0.50\textwidth]{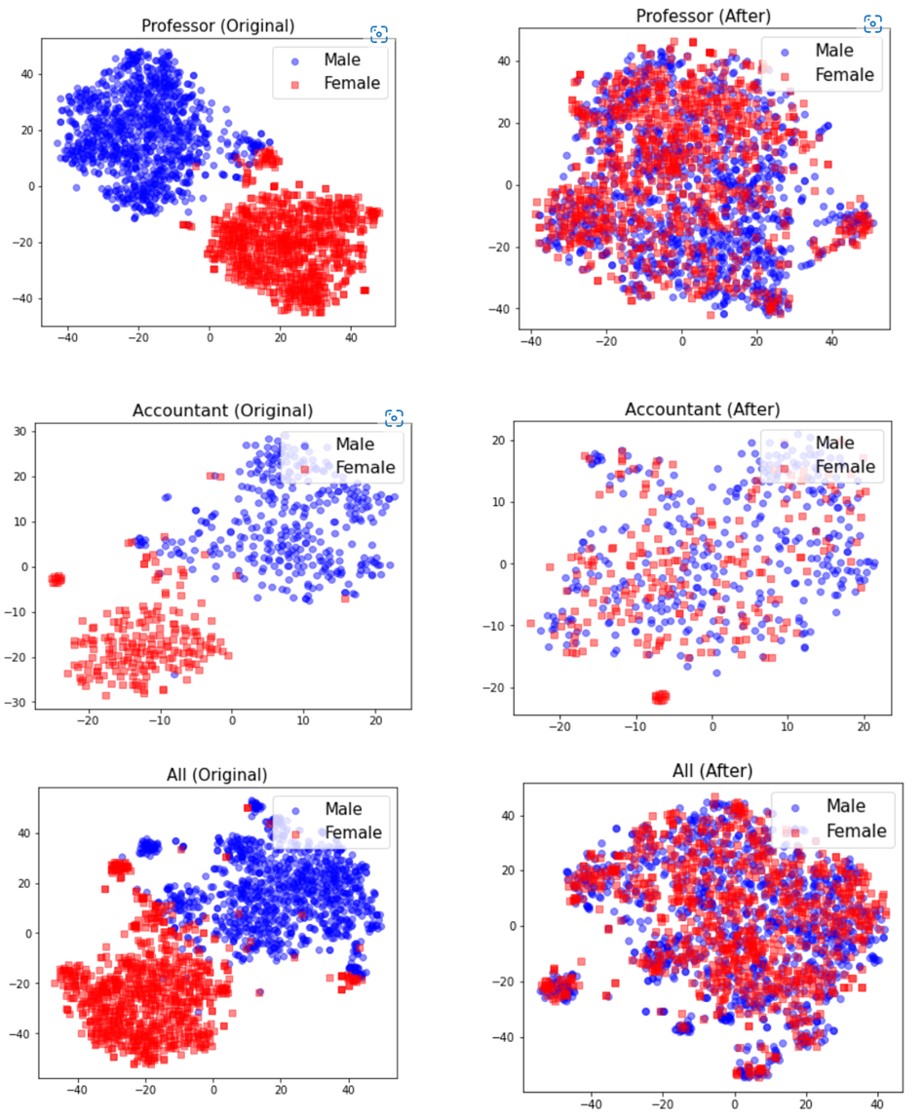}
\vspace{-0.5cm}
\caption{ The t-SNE projections of BERT representations for various professions, including Professors, Accountants, and all proffesions, before and after applying IGBP.}
\label{fig:bios_proj}
\end{figure}
\subsection{Benefits of Non-Linear Information Removal}
\label{app:benefits_nonlin}
\begin{figure}[htb]
\centering

   \includegraphics[width=1\linewidth]{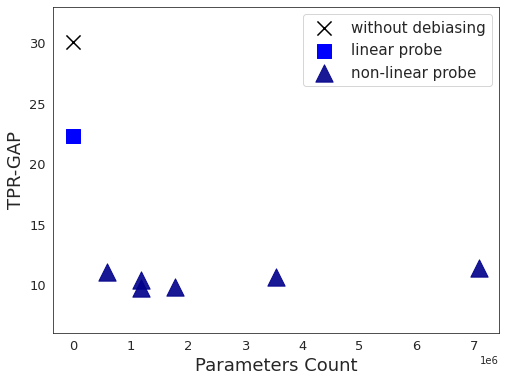}

\caption{The TPR-GAP results obtained by applying \methodName with different probe classifier architectures on the DIAL dataset.}
\label{fig:nonlinear_moji}
\end{figure}

We present the probe architectures we use in our expirement Section \ref{sec:probe_complexity}. These include a linear probe with one layer, and several non-linear probes that use ReLU activations.
From left to right: 
one-hidden layer of the same size as the input dimension, two-hidden layers with the same size as the input dimension, one-hidden layer with size of twice the input dimension, three-hidden layers with size of input dimension, one-hidden layer with size of three times the input dimension.
Figure \ref{fig:nonlinear_moji} shows the  results of Section \ref{sec:probe_complexity} experiment on DIAL dataset. We observe the same trend. The maximum accuracy drop is 1.32\%.
\subsubsection{Number of Iterations}
\begin{figure}[H]
\includegraphics[width=0.48\textwidth,height=0.25\textwidth]{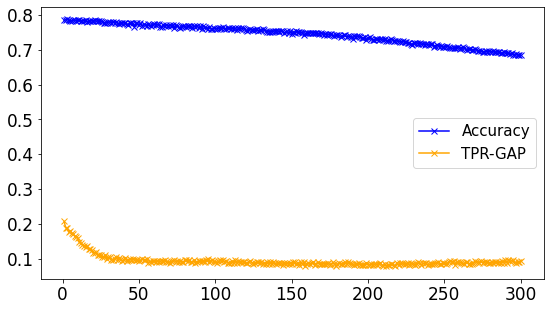}
\vspace{-0.5cm}
\caption{ Mean accuracy and TPR-GAP results versus number of iterations on Bias in Bios dataset with Bert as encoder, averaged over 5 different random seeds. }
\label{fig:bert_iterations}
\end{figure}

We conduct the same experiment of Section \ref{sec:num_it} on DIAL dataset and show the result in Figure \ref{fig:bert_iterations}.
\label{app:bios_it}

\end{document}